\documentclass[11pt,a4paper]{article}
\usepackage[hyperref]{emnlp2018}
\usepackage{times}
\usepackage{latexsym}

\usepackage{url}
\usepackage{adjustbox}
\usepackage{times}
\usepackage{latexsym}
\usepackage{graphicx}
\usepackage{url}
\usepackage{amssymb}
\usepackage{amsmath}
\usepackage{multirow}
\usepackage{booktabs}
\usepackage{subfig}
\usepackage{xcolor}
\usepackage{todonotes}
\usepackage[T1]{fontenc}  
\usepackage{flushend}
\usepackage{etoolbox}

\aclfinalcopy 

\flushend

\title{Open Domain Web Keyphrase Extraction \\ Beyond Language Modeling}

\author{Lee Xiong \quad Chuan Hu \quad Chenyan Xiong  \quad Daniel Campos \quad Arnold Overwijk \\
  Microsoft AI and Research \\
  Redmond, WA 98052, USA \\
  {\tt \{lexion, chuah, cxiong, dacamp, arnoldo\}@microsoft.com}}

\date{}

\begin{document}






\maketitle

\begin{abstract}
This paper studies keyphrase extraction in real-world scenarios where documents are from diverse domains and have variant content quality.
We curate and release OpenKP, a large scale open domain keyphrase extraction dataset with near one hundred thousand web documents and expert keyphrase annotations.
To handle the variations of domain and content quality, we develop BLING-KPE, a neural keyphrase extraction model that goes beyond language understanding using visual presentations of documents and weak supervision from search queries.
Experimental results on OpenKP confirm the effectiveness of BLING-KPE and the contributions of its neural architecture, visual features, and search log weak supervision.
Zero-shot evaluations on DUC-2001 demonstrate the improved generalization ability of learning from the open domain data compared to a specific domain.
\end{abstract}
\section{Introduction}

Automatically extracting keyphrases that are salient to the document meanings is an essential step to semantic document understanding. An effective keyphrase extraction (KPE) system can benefit a wide range of natural language processing and information retrieval tasks~\cite{turney2001mining, hasan2014automatic}.
Recent neural methods formulate the task as a document-to-keyphrase sequence-to-sequence task. These neural KPE models have shown promising results compared to previous systems~\cite{chen2018keyphrase, meng2017deep, ye2018semi}.

Noticeably, the recent progress in neural KPE is mostly observed in documents originating from the scientific domain~\cite{meng2017deep, augenstein2017semeval}.
Perhaps because the scientific domain has sufficient training data for these neural methods: Authors are in the practice of assigning keyphrases to their publications.  
In real-world scenarios, most potential applications of KPE deal with diverse documents originating from sparse sources that are rather different from scientific papers.
They often include a much diverse document structure and reside in various domains whose contents target much wider audiences than scientists. 
It is unclear how well the neural methods trained in the scientific domain generalize to other domains and in real-world scenarios.

This paper focuses on the task of open domain web keyphrase extraction, which targets KPE for web documents without any restriction of the domain, quality, nor content of the documents.
We curate and release a large scale open domain KPE dataset, OpenKP, which includes about one hundred thousand web documents with expert keyphrase annotations.\footnote{The dataset, resources, and future updates are available at \url{aka.ms/BLING}.}
The web documents are randomly sampled from the English fraction of a large web corpus and reflect the characteristics of typical web pages, with large variation in their domains and content qualities.
To the best of our knowledge, this will be the first publicly available open domain manually annotated keyphrase extraction dataset at this scale.


This paper develops BLING-KPE, \textbf{B}eyond \textbf{L}anguage Understand\textbf{ING} \textbf{K}ey\textbf{P}hrase \textbf{E}xtraction, that tackles the challenges of KPE in documents from variant domains and content qualities.
BLING-KPE uses a convolutional transformer architecture to model the language properties in the document, while also goes beyond by introducing the visual representation of the document and weak supervision from search user clicks.

The \emph{visual presentations} of the document, including the location, size, font, and HTML structure of each text piece in the document, are integrated as visual features to the word embeddings in BLING-KPE. 
BLING-KPE learns to model the visual representations together with the document language in its network.

The \emph{weak supervision from search clicks} is formulated as a pre-training task: Query Prediction.
It trains the model to predict which phrase in the document has been used as a ``click query'', a query that a user issued to search and click on the document.
The click queries on a document reflect the user's perceptions of the relatedness and importance when searching the document and can be considered as pseudo keyphrases.
Pre-training on this weak supervision brings in training signals available at scale in commercial search systems.

Our experiments on OpenKP demonstrate the effectiveness of BLING-KPE.
It outperforms standard KPE baselines, recent neural approaches and a highly optimized commercial KPE system by large margins.
Ablation studies show the contributions of the neural architecture, visual features, and search weak supervision to BLING-KPE; removing any of them significantly reduces its accuracy.

Another advantage of learning from real-world open domain documents is improved generalization ability.
We conduct zero-shot evaluations on the DUC-2001 news KPE datasets~\cite{wan2008single}, where neural KPE systems are evaluated without seeing any labels from their news articles.
BLING-KPE trained on OpenKP is the only neural method that outperforms traditional non-neural KPE methods, while neural KPE systems trained on the scientific documents do not generalize well to the news domain due to the domain differences.

\section{Related Work}

The classic keyphrase extraction systems typically include two components: \emph{candidate keyphrase extraction} and \emph{keyphrase importance estimation}~\cite{hasan2014automatic}.
The candidate keyphrases are often extracted by heuristic rules, for example, finding phrases following certain POS tag sequences~\cite{wan2008single, liu2009unsupervised, mihalcea2004textrank}, pre-defined lexical patterns~\cite{nguyen2009ontology, medelyan2009human}, or using entities as candidate phrases~\cite{grineva2009extracting}.

The importance of the candidate keyphrases can be estimated by unsupervised or supervised methods. The unsupervised methods leverage the graph structures between phrases in the document~\cite{mihalcea2004textrank, wan2008collabrank, wan2008single}, and topic information from topic modeling~\cite{grineva2009extracting, liu2009clustering, liu2010automatic}.
The supervised keyphrase selection methods formulate a classification or ranking task and combine features from phrase frequencies~\cite{witten2005kea}, document structures~\cite{chen2005practical, yih2006finding}, and external resources such as Wikipedia~\cite{medelyan2009human} and query log~\cite{yih2006finding}.

Recently, neural techniques have been applied to keyphrase tasks.
Meng et al. formulate a seq2seq learning task that learns to extract and generate the keyphrase sequence from the document sequence; they incorporate a copy mechanism to the seq2seq RNN to extract phrases in the generation process (CopyRNN)~\cite{meng2017deep}. Improving this seq2seq setup has been the focus of recent research, for example, adding diverse constraints to reduce the duplication of produced keyphrases~\cite{yuan2018generating, chen2018keyphrase}, bringing auxiliary tasks to reduce the needs of training data~\cite{ye2018semi}, and adding title information to improve model accuracy~\cite{chen2018guided}.

The recent neural KPE methods have shown strong performances on the scientific domain, where large scale training data is available from the author assigned keyphrases on papers~\cite{meng2017deep}. Such specific domain training data limits the model generalization ability. Chen et al. show the seq2seq keyphrase generation models trained on scientific papers do not generalize well to another domain~\cite{chen2018keyphrase}.

In general, previous research finds automatic keyphrase extraction a challenging task: its state-of-the-art accuracy is much lower than other language processing tasks, while supervised methods do not necessarily outperform simple unsupervised ones. \citet{hasan2014automatic} pointed out potential ways to improve automatic keyphrase extraction, including better incorporation of \textit{background knowledge}, better handling \textit{long documents}, and better \textit{evaluation} schemes.
BLING-KPE aims to address these challenges by incorporating pre-training as a form of \textit{background knowledge}, visual information to improve \textit{long document} modeling, and OpenKP as a large scale open domain \textit{evaluation} benchmark.


\section{Open Domain Keyphrase Benchmark}
\label{sec:data}
This section describes the curation of \texttt{OpenKP} and its notable characteristics. 

\subsection{Data Curation}
\noindent
\textbf{Documents} in \texttt{OpenKP} include about seventy thousand web pages sampled from the index of Bing search engine.\footnote{A new OpenKP version with 150K documents is available at \url{msmarco.org}.} The sampling is conducted on the pool of pages seen by United State users between Nov 2018 and Feb 2019.

There is no restriction on the domain or type of documents. 
They can be content-oriented pages like news articles, multi-media pages from video sites, or indexing pages with many hyperlinks. \texttt{OpenKP} is designed to reflect the diverse properties of web documents in the internet.

\textbf{Keyphrase Labels} are generated by our expert annotators.
For each document, they examine the rendered web page and manually label 1-3 keyphrases following these definitions:
\begin{itemize}\setlength\itemsep{0em}
    \item \textit{Salience:} A keyphrase captures the essential meaning of the page with no ambiguity.
    \item \textit{Extraction:} The keyphrase has to appear in the document.
    \item \textit{Fine-Grained:} The keyphrase cannot be general topics, such as ``Sports'' and ``Politics''.
    \item \textit{Correct \& Succinct:} The keyphrase has to form a correct English noun phrase, while also cannot be clauses or sentences.
\end{itemize}
We use the extraction setting to ensure labeling consistency and to increase annotation speed, which is around 42 pages per hour.

\textbf{Expert Agreements.}  Our annotation experts are trained employees dedicated to providing high-quality annotations on web documents.
We follow standard practice in generating annotations for production systems, which included regular touchpoints to understand the confusion, as well as updates on the judgment guidelines to resolve ambiguities.

To study the task difficulty, we had five judges each annotate the same 50 random URLs.
We measure the pairwise agreements between experts at different depths by Exact Match on the whole keyphrase, as well as the overlap between select keyphrases' unigrams.
The agreement between judges is listed in Table~\ref{tab:agreements}.

The results confirm that open domain keyphrase extraction is not an easy task. When measuring agreement for the top 3 keyphrases, our expert judges completely agree on about $43\%$ of keyphrase pairs.
Compared to the previous small scale annotations, for example, on DUC-2001's news articles~\cite{wan2008single}, annotating web pages with diverse contents and domains are harder.

We manually examined these annotations and found two sources of disagreement: Chunking Variances and KP Choices. 

\textit{Chunking Variances} we define as two judges pick different boundaries of the same concept. For example, one judge may select ``Protein Synthesis'' as the keyphrase, and others may select ``Protein'' and ``Synthesis'' as two separate keyphrases. We found Chunking Variances consist of about 20\% of disagreements.
As shown in Table~\ref{tab:agreements}, the judge agreements is substantially higher on Unigram overlaps than on Exact matches, indicating that they may select chunks that overlap with each other but not exactly the same.

\textit{KP Choices} we define as two judges pick different keyphrases. The judges agree mostly (64\%) on the first entered keyphrase, as shown in Table~\ref{tab:agreements}. 
The variations on the second and third keyphrases are larger. However, we found the variations are more about which keyphrases they choose to enter, \emph{not about whether a phrase is a keyphrase or not}.
\begin{table}[t]
    \centering
    \caption{The agreements between pairs of expert judges at different annotation depth. Exact and Unigram show the percentage of judge agreement on exact keyphrases and overlapped unigrams. 
    \label{tab:agreements}} 
    \small
    \begin{tabular}{l|c|c} \hline \hline
    \textbf{Judge Depth} 
    & \textbf{Exact Match}  & \textbf{Unigram Match} \\ \hline
    Keyphrase@1 & 64.74\% & 64.74\% \\ \hline
    Keyphrase@2 & 48.30\% & 63.12\% \\ \hline
    Keyphrase@3 & 43.51\% & 57.66\% \\ \hline \hline
    \end{tabular}
\end{table}

The variations on judge labels mostly reflect the missing positives in \texttt{OpenKP}; most of the keyphrases annotated by judges are correct. 
We can reduce the missing positives by a deeper annotation, i.e. ten keyphrases per document, or by labeling all candidate phrases with classification labels~\cite{liu2018automatic}. However, that will significantly reduce the number of total documents in \texttt{OpenKP}, as each document costs much more to annotate.
We chose the current design choice of \texttt{OpenKP} to favor a larger amount of training labels, which, in our experience, is more effective in training deep neural models.

\begin{figure}[t]
    \centering
    \includegraphics[width=0.95\columnwidth]{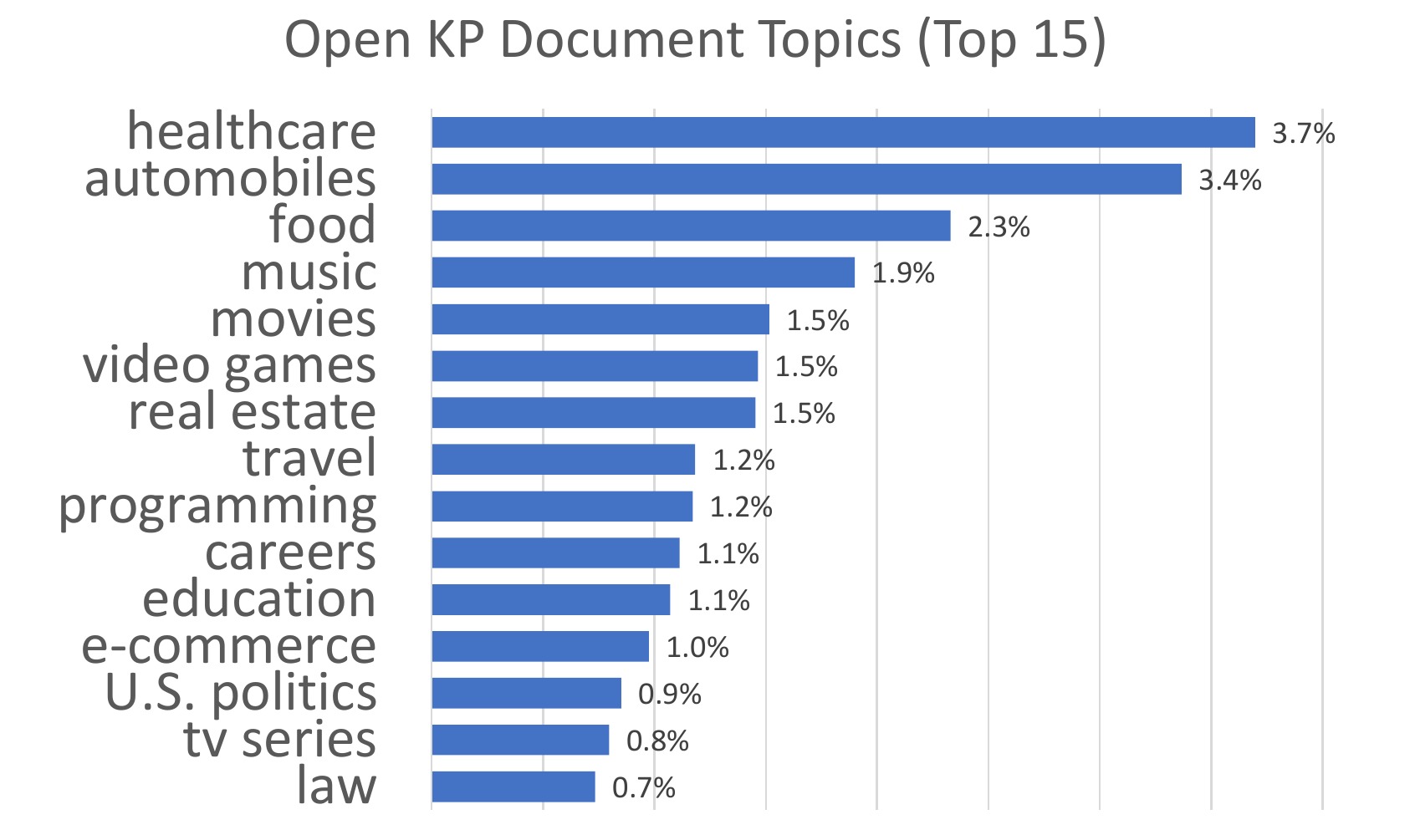}
    \caption{The most popular topics in OpenKP .
    \label{fig:domain}}
\end{figure}

\subsection{Data Characteristics}

\begin{table}[t]
    \centering
    \caption{Statistics of OpenKP used in our experiments. The new version on MSMARCO include 150K documents.
    \label{tab:stat}} \small
    \begin{tabular}{l|c|c} \hline \hline
      \textbf{Statistics} & \textbf{Mean} & \textbf{STD}  \\ \hline
    
    Doc Length & 900.4 & 1494.4  \\
    \# of KPs per Doc & 1.8 & 0.8 \\ 
    Keyphrase Length & 2.0 & 0.96 \\ \hline
    Doc Vocabulary Size & 1.5M & n.a. \\
    KP Vocabulary Size & 62K & n.a. \\
    \hline
    \# of Documents & 68K & n.a. \\
    \# of Unique KPs & 99.6K & n.a. \\ 
    \hline \hline
    \end{tabular}
\end{table}

Table~\ref{tab:stat} lists the statistics of \texttt{OpenKP}.
The document length is the length of the text parsed from the HTML of the web page, using a production HTML parser. The parsed texts will be released with the dataset. These statistics reflect the large variations in the document contents; their length varies a lot and share little common keyphrases, as shown by a large number of unique keyphrases.

We also leverage a production classifier to classify \texttt{OpenKP} documents into 5K predefined domains. The top 15 most popular classes and their distributions are shown in Figure~\ref{fig:domain}.
As expected, these documents have a large variation in their topic domains. The most popular domain, ``healthcare'', only covers $3.7\%$ documents; the tenth most popular topic only covers 1\% of documents. Moreover, the top 15 classes make up less than 25\% of the entire dataset which showcases what a domain diverse dataset \texttt{OpenKP} is.


\section{Keyphrase Extraction Model}
\label{sec:model}
This section describes the architecture, visual features, and weak supervision of BLING-KPE.

\subsection{Network Architecture}
\label{sec:network}

\begin{figure}[t]
    \centering
    \includegraphics[width=0.95\columnwidth]{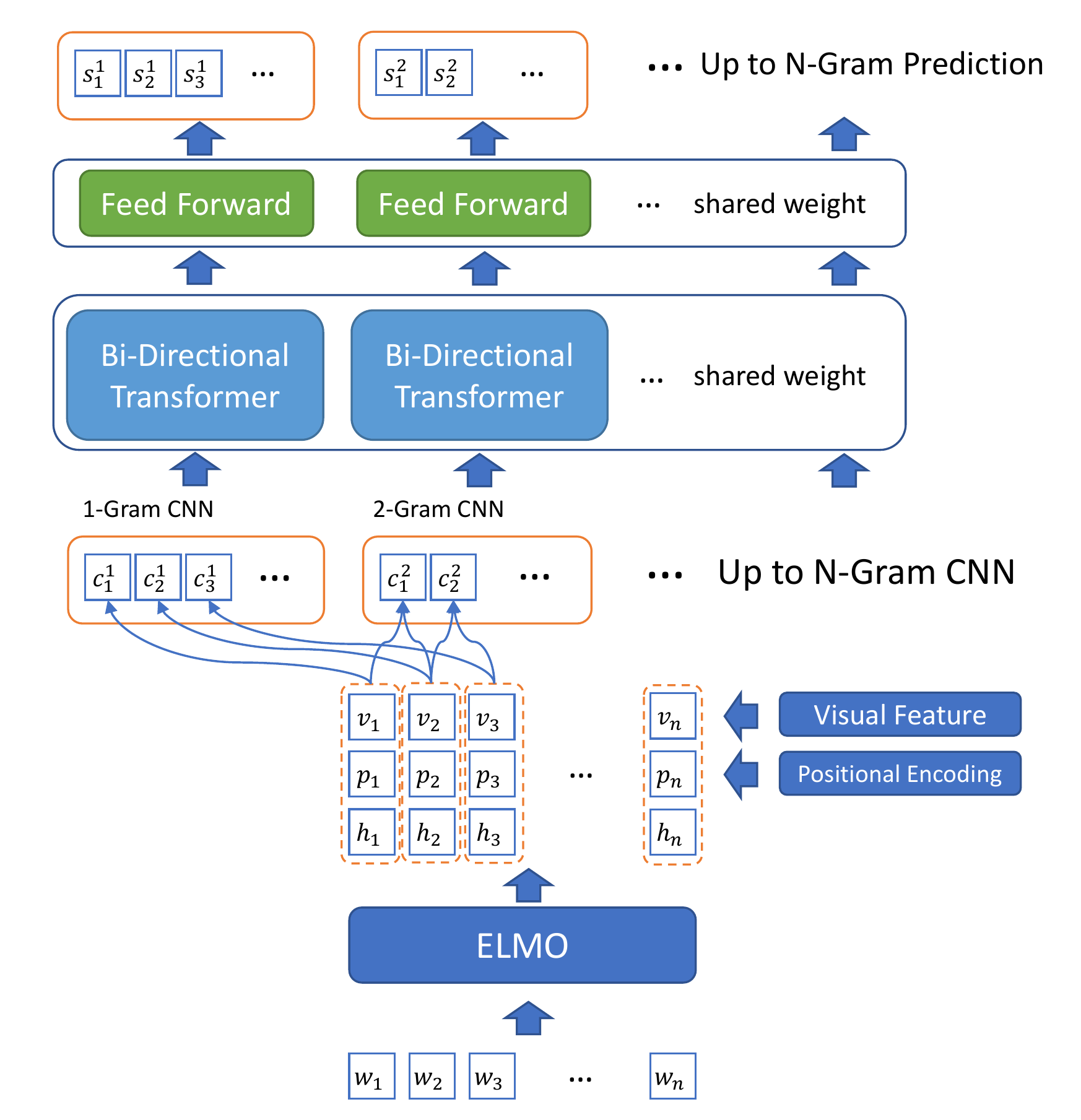}
    \caption{The BLING-KPE model architecture}
    \label{fig:model}
\end{figure}

As shown in Figure~\ref{fig:model}, BLING-KPE is a keyphrase \emph{extraction} model.
It takes the word sequence of the document, $d=\{w_1,...w_i,...w_n\}$, and assigns keyphrase scores to its n-gram: $f(w_{i:i+k}, d)$. 
This process includes two main components: Hybrid Word Embedding and Convolutional Transformer.

\textbf{Hybrid Word Embedding.} BLING-KPE represents each word by its ELMo embedding, position embedding, and visual features.

The ELMo embedding brings the local contextual information:
\begin{align}
    \vec{h}_i &= \text{ELMo}(w_i).
\end{align}
The standard pre-trained ELMo is used~\cite{peters20018elmo}.

The position embedding models the location the word in the document content. 
It uses the standard sinusoidal position embedding~\cite{transformer}:
\begin{align}
    \vec{\text{pos}}_i (2p) &= sin(i/10000^{2p/P}), \\
    \vec{\text{pos}}_i (2p+1) &= cos(i/10000^{2p/P}).
\end{align}
The p-th dimension of the position embedding is a function of its position (i) and dimension (p). 

The visual features represent the visual presentation of each word. We denote the visual feature as $\vec{v}_i$ and will describe its details in $\mathsection$\ref{sec:visual}.

The hybrid word embedding is the concatenation of the three:
\begin{align}
    \vec{w_i} &= \vec{h}_i \frown \vec{\text{pos}}_i \frown \vec{v}_i.
\end{align}

\textbf{Convolutional Transformer.} BLING-KPE uses a convolutional transformer architecture
to model n-grams and their interactions.

It first composes the hybrid word embeddings to n-gram embeddings using CNNs. The embedding of i-th k-gram is calculated as
\begin{align}
    \vec{g}_i^k &= \text{CNN}^k(\vec{w}_{i:i+k}),
\end{align}
where $k$ is the length of the n-gram, $1\leq k \leq K$. $K$ is the maximum length of allowed candidate n-grams. Each k-gram has its own set of convolution filters $\text{CNN}^k$ with window size $k$ and stride 1.

It then models the interactions between k-grams using Transformer~\cite{transformer}.
\begin{align}
    \vec{t}_i^k &= \text{Transformer}_i(\vec{G}^k), \\
    \vec{G}^k &= \vec{g}_1^k \frown ... \frown \vec{g}_i^k ... \frown \vec{g}_{n-k+1}^k.
\end{align}
The sequence $\vec{G}^k$  is the concatenations of all k-gram embeddings.
The Transformer models the self-attentions between k-grams and fuses them to global contextualized embeddings.

The Transformer is convolutional on all length $k$ of n-grams;
the same parameters are used model the interactions between n-grams at each length, to reduce the parameter size. The intuition is that the interactions between bi-grams and that between tri-grams are not significantly different.

The final score of an n-gram is calculated by a feedforward layer upon the Transformer. Like the Transformer, the same feedforward layer is applied (convolutional) on all n-grams.
\begin{align}
    f(w_{i:i+k}, d) &= \text{softmax}_{i, k} (s_i^k),\\
        s_i^k &= \text{Feedforward}(\vec{t}_i^k).
\end{align}
The softmax is taken over all possible n-grams at each position $i$ and each length $k$. The model decides the span location and length jointly.

\textbf{Learning.} The whole model is trained as a classification problem using cross-entropy loss:
\begin{align}
    l = \text{Cross-Entropy}(y_i^k, f(w_{i:i+k},d)),
\end{align}
where $y_i^k$ is the label of whether the phrase $w_{i:i+k}$ is a keyphrase of the document.

\subsection{Visual Features}
\label{sec:visual}

\begin{table}[t]
    \centering
     \caption{Visual Features. All features are extracted at per word level and the parent block level (the parent node of the word in the HTML DOM tree).
    \label{tab:visual}} \small
    \begin{tabular}{l|c} \hline \hline
        \textbf{Name} & \textbf{Dimension} \\ \hline
        Font Size &  1$\times$2 \\
        Text Block Size  & 2$\times$2 \\ \hline
        Location in Rendered Page  & 2$\times$2 \\ \hline
        Is Bold Font  &  1$\times$2 \\  \hline
        Appear In Inline  &  1$\times$2 \\ 
        Appear In Block  &  1$\times$2 \\ 
        Appear In DOM Tree Leaf  &  1$\times$2 \\ 
        \hline \hline
    \end{tabular}
   
\end{table}
We extract four groups of visual features for each word in the document.
\begin{itemize}\setlength\itemsep{0em}
    \item \textbf{Size} features include the height and width of the text block a word appears in.
    \item \textbf{Location} features include the 2-d location of the word in the rendered web page.
    \item \textbf{Font} feature includes the font size and whether the word is in Bold.
    \item \textbf{DOM} features include whether the word appears in ``inline'' or ``block'' HTML tags, also whether it is in a leaf node of the DOM tree.
\end{itemize}
The full feature set is listed in Table~\ref{tab:visual}. We double the features by including the same features from the word's parent block in the DOM tree. The visual features are included in the OpenKP releases.

\subsection{Weak Supervisions from Search}
\label{sec:training}

An application of keyphrases is information retrieval. The extracted keyphrases are expected to capture the main topic of the document, thus can provide high quality document indexing terms~\cite{gutwin1999improving} or new semantic ranking features~\cite{xiong2018towards}. 
Reversely, user clicks bring the user's perception of the document during the search and provide a large number of feedback signals for document understanding~\cite{croft2010search}.

BLING-KPE leverages the user feedback signals as weak supervision, in the task of Query Prediction.
Given the document $d$, BLING-KPE learns to predict its click queries $Q=\{q_1,...,q_m\}$. 

This pre-training step uses the same cross entropy loss:
\begin{align}
    l_{pre} & = \text{Cross-Entropy} (y'_i, f(q_i,d)),
\end{align}
where $y'_i$ indicates whether the query $q_i$ is a click query and also appears as an n-gram in the document $d$.
The Query Prediction labels exist at scale in commercial search logs and provide a large number of pre-training signals.

\begin{table}[t]
    \centering
        \caption{Statistics of Query Prediction Dataset. The data is from a sample of Bing search log in on week.
    \label{tab:qpre}}
    \small
    \begin{tabular}{l|c|c} \hline \hline
      \textbf{Statistics}   & \textbf{Mean} & \textbf{STD}  \\ \hline
    Doc Length & 1211.8 & 1872.6 \\
    \# of Query per Doc & 1.32 & 0.76 \\
    Query Length & 2.4 & 1.07 \\ \hline
    Doc Vocabulary Size & 15M & n.a \\ 
    Query Vocabulary Size & 383K & n.a \\ \hline
    \# of Documents & 1.6M & n.a. \\
    \# of Unique Queries & 1.5M & n.a. \\ \hline \hline
    \end{tabular}

\end{table}
\section{Experimental Methodology}
\label{sec:exp}

\indent
\textbf{Datasets} used in our experiments include  OpenKP, as described in $\mathsection$\ref{sec:data}, Query Prediction, and DUC-2001~\cite{wan2008single}.

The Query Prediction data is sampled from the Bing search log with navigational and offensive queries filtered out. We keep only the click queries that are included as an n-gram in the document to be consistent with OpenKP's extractive setting.
The statistics of the sample is listed in Table~\ref{tab:qpre}.

DUC-2001 is the KPE extraction dataset on DUC news articles~\cite{wan2008single}. It includes 309 news articles and on average 8 keyphrase per article.

We use random 80\%-20\% train-test splits on OpenKP and Query Prediction.
On OpenKP, BLING-KPE is first pre-trained on Query Prediction and then further trained on its manual labels. There is no overlap between the documents in Query Prediction and OpenKP.


DUC-2001 uses the zero-shot evaluation setting from prior research~\cite{meng2017deep, chen2018keyphrase}; no labels in DUC-2001 are used to train nor validate the neural models. It tests neural models' generalization ability from the training domain to a different testing domain.

\textbf{Evaluation Metrics.} OpenKP and Query Prediction use \textbf{P}recision and \textbf{R}ecall@\{1, 3, 5\}. DUC-2001 uses F1@10, the same as prior research~\cite{meng2017deep, chen2018keyphrase}.

Statistically, significant improvements are evaluated by permutation test with p$<$0.05 on OpenKP and Query Prediction. The baselines on DUC-2001 reuse scores from previous results; the statistical significant test is not applicable as per document results are not shared.

\textbf{Baselines.} OpenKP and Query Prediction experiments compare \texttt{BLING-KPE} with: traditional KPE methods, production systems, and a neural baseline.

Traditional KPE baselines include the follows.
\begin{itemize}\setlength\itemsep{0em}
    \item {TFIDF} is the unsupervised frequency based KPE system. The IDF scores are calculated on the corresponding corpus.
    \item {TextRank} is the popular graph-based unsupervised KPE model~\cite{mihalcea2004textrank}. Our in-house implementation is used.
    \item {LeToR} is the feature-based KPE model. It use LambdaMart~\cite{Burges2010FromRT} and standard KPE features, i.e. those in KEA~\cite{witten2005kea}.
\end{itemize}

The production baselines include two versions.
\begin{itemize}\setlength\itemsep{0em}
    \item {PROD} is our current feature-based production KPE system. It uses many carefully engineered features and LambdaMart.
    \item {PROD (Body)} is the same system but only uses the body text, i.e. the title is not used.
\end{itemize}

All these unsupervised and feature-based methods use the same keyphrase candidate selection system with PROD.

\begin{table}[t]
    \centering
        \caption{Parameters to learn in BLING-KPE.
    \label{tab:para}}
    \small
    \begin{adjustbox}{width=0.95\columnwidth,center}
    \begin{tabular}{l|c} \hline \hline
    \textbf{Component} & \textbf{Dimensions} \\ \hline
     ELMo    &  pre-trained and frozen \\
     Position Embedding & 256 \\
     Visual Feature & 18 \\\hline
     N-gram CNN & 512 filter, 1-5 window size (ngram) \\
     Transformer & 8 head, 512 hidden dimension\\ 
     Feedforward & 512-relu-512-relu-1 \\ \hline
    \end{tabular}
    \end{adjustbox}
\end{table}

The neural baseline is CopyRNN~\cite{meng2017deep}. We use their open-source implementation and focus on the OpenKP dataset which is publicly available.

\textbf{Implementation Details.} Table~\ref{tab:para} lists BLING-KPE parameters. 
The training uses Adam optimizer, learning rate 0.3 with logarithmic decreasing to 0.001,  batch size 16, and 0.2 dropout probability in n-gram CNN, Transformer and feedforward layers.
Learning takes about 2.5 hours (2 epochs) to converge on \texttt{OpenKPE} and about 13 hours (3 epochs) on Query Prediction, based on validation loss. In BLING-KPE, the maximum document length is 256  and documents are zero-padded or truncated to this length. Baselines use the original documents, except CopyRNN which works better with 256. 
The maximum n-gram length is set to five (K=5).

\begin{table*}[ht]
    \centering
     \caption{Keyphrase Extraction Accuracy. 
     \textbf{Bold} marks statistically significant improvements over all baselines.
    \label{tab:overal}}
    \small
    \begin{adjustbox}{width=0.95\textwidth,center}
    \begin{tabular}{l|cc|cc|cc|cc|cc|cc} \hline \hline
    & \multicolumn{6}{c|}{\textbf{OpenKP}} 
    & \multicolumn{6}{c}{\textbf{Query Prediction}} \\ \hline  
           \textbf{Method}     
    &  \textbf{P@1} &  \textbf{R@1}
     &  \textbf{P@3} &  \textbf{R@3}
      &  \textbf{P@5} &  \textbf{R@5} 
      &  \textbf{P@1} &  \textbf{R@1}
     &  \textbf{P@3} &  \textbf{R@3}
      &  \textbf{P@5} &  \textbf{R@5} 
    \\ \hline
{TFIDF} 
& 0.283 & 0.150 & 0.184 & 0.284 & 0.137 & 0.347  & 0.403 & 0.332 & 0.204 & 0.491 & 0.133 & 0.526
\\ 
{TextRank} 
& 0.077 & 0.041 & 0.062 & 0.098 & 0.055 & 0.142 & 0.132 & 0.111 & 0.089 & 0.218 & 0.073 & 0.295 \\ 
{LeToR} 
& 0.301 & 0.158 & 0.173 & 0.268 & 0.127 & 0.324 & 0.328 & 0.271 & 0.169 & 0.406 & 0.119 & 0.471 \\ \hline
{PROD}
& 0.353 & 0.188 & 0.195 & 0.299 & 0.131 & 0.331 & 0.376 & 0.308 & 0.197 & 0.468 & 0.129 & 0.505 \\ 
{PROD (Body)}
& 0.214 & 0.094 & 0.130 & 0.196 & 0.094 & 0.234 & 0.353 & 0.287 & 0.191 & 0.454 & 0.125 & 0.492 \\
\hline
{CopyRNN}
& 0.288 & 0.174 & 0.185 & 0.331 & 0.141 & 0.413 & -- & -- & -- & -- & -- & -- \\ \hline
{BLING-KPE} & \textbf{0.404} & \textbf{0.220} & \textbf{0.248} & \textbf{0.390} & \textbf{0.188} & \textbf{0.481} & \textbf{0.540} & \textbf{0.449} & \textbf{0.275} & \textbf{0.654} & \textbf{0.188} & \textbf{0.729} \\ \hline \hline
    \end{tabular}
   \end{adjustbox}
\end{table*}

\begin{table*}[h]

    \centering
     \caption{Performance of BLING-KPE ablations. 
     \textit{Italic} marks statistically significant worse performances than Full Model.
    \label{tab:ablation}}
    \begin{adjustbox}{width=0.95\textwidth,center}
    \small
    \begin{tabular}{l|cc|cc|cc|cc|cc|cc} \hline \hline
    & \multicolumn{6}{c|}{\textbf{OpenKP}} 
    & \multicolumn{6}{c}{\textbf{Query Prediction}} \\ \hline  
       \textbf{Method}     
    &  \textbf{P@1} &  \textbf{R@1}
     &  \textbf{P@3} &  \textbf{R@3}
      &  \textbf{P@5} &  \textbf{R@5} 
      &  \textbf{P@1} &  \textbf{R@1}
     &  \textbf{P@3} &  \textbf{R@3}
      &  \textbf{P@5} &  \textbf{R@5} 
    \\ \hline
{No ELMo} 
& \textit{0.270} & \textit{0.145} & \textit{0.172} & \textit{0.271} & \textit{0.132} & \textit{0.347} & \textit{0.323} & \textit{0.274} & \textit{0.189} & \textit{0.450} & \textit{0.136} & \textit{0.527} \\
{No Transformer} 
& \textit{0.389} & \textit{0.211} & 0.247 & \textit{0.385} & 0.189 & 0.481 & \textit{0.489} & \textit{0.407} & \textit{0.258} & \textit{0.618} & \textit{0.178} & \textit{0.698} \\
{No Position} 
& \textit{0.394} & \textit{0.213}  & 0.247 & \textit{0.386} &  0.187 & \textit{0.475} & 0.543 & 0.452 & 0.281 & 0.666  & 0.191 & 0.742 \\ \hline
{No Visual} 
& \textit{0.370} & \textit{0.201}  & \textit{0.230} & \textit{0.362} & \textit{0.176} & \textit{0.450}  & \textit{0.492} & \textit{0.409} & \textit{0.258} & \textit{0.615}  & \textit{0.178} & \textit{0.695} \\    
{No Pretraining} 
& \textit{0.369} & \textit{0.198}  & \textit{0.236} & \textit{0.367} & \textit{0.181} & \textit{0.460}  
& -- & -- &-- & -- & -- &--
\\ \hline
{Full Model}
& 0.404 & 0.220 & 0.248 & 0.390 & 0.188 & 0.481 & 0.540 & 0.449 & 0.275 & 0.654 & 0.188 & 0.729 \\ 
\hline \hline
\end{tabular}
\end{adjustbox}
\end{table*}

\section{Evaluation Results}
\label{sec:eva}
Three experiments are conducted to evaluate the accuracy of BLING-KPE, the source of its effectiveness, and its generalization ability.

\subsection{Overall Accuracy}
The overall extraction accuracy on OpenKP and Query Prediction is shown in Table~\ref{tab:overal}.

TFIDF works well on both tasks.
Frequency-based methods are often strong baselines in document representation tasks. 
LeToR performs better than its frequency feature {TFIDF} in {OpenKP} but worse on Query Prediction. Supervised methods are not necessarily stronger than unsupervised ones in KPE~\cite{hasan2014automatic}. TextRank does not work well in our dataset; its word graph is likely misguided by the noisy contents.

PROD, our feature-based production system, outperforms all other baselines by large margins on OpenKP. It is expected as it is highly optimized with a lot of engineering efforts. 
Nonetheless, adapting a complex feature-based system to a new task/domain requires extra engineering work; directly applying it to the Query Prediction task does not work well.
The feature-based Production system also needs the title information; PROD (Body) performs much worse than PROD. 

CopyRNN performs relatively well on OpenKP, especially on later keyphases.
The main challenge for CopyRNN is the low-quality and highly variant contents on the web.
Real-world web pages are not cohesive nor well-written articles but include various structures such as lists, media captions, and text fragments.
Modeling them as a word sequence is not ideal.
The other differences are not as significant:
The vocabulary size and training data size on Query Prediction are similar to CopyRNN's KP 40K dataset;
CopyRNN performs better on keyphrase extraction than generation in {KP20k}~\cite{meng2017deep}.

BLING-KPE outperforms all other methods by large margins. The improvements are robust and significant on both tasks, both metrics, and on all depths.
It achieves 0.404 P@1 on OpenKP and recovers 72\% of clicked queries at depth 5 on Query Prediction. The sources of this effectiveness is studied in the next experiment.

\subsection{Ablation Study}
Table~\ref{tab:ablation} shows ablation results on BLING-KPE's variations.
Each variation removes a component and keeps all others unchanged.

\textbf{ELMo Embedding.} We first verify the effectiveness of using ELMo embedding by replacing ELMo with the WordPiece token embedding~\cite{DBLP:journals/corr/WuSCLNMKCGMKSJL16}.
The accuracy of this variation is much lower than the accuracy of the full model and others. The result is shown in the first row of Table~\ref{tab:ablation}. 
The context-aware word embedding is a necessary component of \texttt{BLING-KPE}. 

\textbf{Network Architecture.} 
The second part of Table~\ref{tab:ablation} studies the contribution of Transformer and position embedding.
Transformer contributes significantly to Query Prediction; with a lot of training data, the self-attention layers capture the global contexts between n-grams. But on OpenKP, its effectiveness is mostly observed on the first position.
The position embedding barely helps, since real-world web pages are often not one text sequence.

\textbf{Beyond Language Understanding.} As shown in the second part of Table~\ref{tab:ablation}, both visual features and search pretraining contribute significantly to BLING-KPE's effectiveness. Without either of them, the accuracy drops significantly.
Visual features even help on Query Prediction, though users issued the click queries and clicked on the documents \textit{before} seeing its full page. 

The crucial role of ELMo embeddings confirm the benefits of bringing background knowledge and general language understanding, in the format of pre-trained contextual embedding, in keyphrase extraction.
The importance of visual features and search weak supervisions confirms the benefits of going beyond language understanding in modeling real-world web documents.

\begin{table}[t]
    \centering
        \caption{Performance on DUC-2001. Neural models are evaluated directly on DUC-2001 without fine-tuning on DUC labels. Results better than TFIDF are marked \textbf{Bold}.
    \label{tab:duc}}
    \small 
    \begin{tabular}{c|c|c|c} \hline \hline
\textbf{Method}         &  \textbf{F1@10}  
& \textbf{Method}         &  \textbf{F1@10} \\ \hline
TFIDF         & 0.270
& TopicRank & 0.154 \\
TextRank & 0.097 
& KeyCluster & 0.140 \\ 
SingleRank & 0.256 
& ExpandRank & 0.269 \\\hline
\multicolumn{4}{c}{\textbf{Trained by Scientific Papers}} \\ \hline
CopyRNN & 0.164 & CorrRNN & 0.173 \\ \hline
\multicolumn{3}{l|}{BLING-KPE (No Visual, No Pretraining)} &  0.267 \\ \hline 
\multicolumn{4}{c}{\textbf{Trained by Open Domain Documents}} \\ \hline
\multicolumn{3}{l|}{BLING-KPE (No Visual, No Pretraining)} & \textbf{0.282} \\
\hline \hline 
    \end{tabular}

\end{table}
\subsection{Generalization Ability}
\label{sec:gen}
This experiment studies the generalization ability of BLING-KPE using the zero-shot evaluation on DUC-2001~\cite{meng2017deep, chen2018keyphrase}.
For fair comparisons, we use KP20K or OpenKP, the two public datasets, to train the No Visual \& No Pretraining version of BLING-KPE, and evaluate on DUC-2001 directly. No labels in DUC are used to fine-tune the neural models.

To adjust to DUC's larger number of keyphrases, we apply the trained BLING-KPE on the 256-length  chunks of DUC articles and merge the extracted keyphrases using simple heuristics:
\begin{itemize}
\setlength\itemsep{0em}
    \item Weighted Sum: Scores of the same keyphrase from different chunks are summed with weights 0.9$^{p}$. P is the index of the chunk.
    \item Deduplication: A keyphrase is discarded if it is a sub-string of a top 1/4 ranked keyphrase.
\end{itemize}

The results are shown in Table~\ref{tab:duc}. \texttt{BLING-KPE}, when trained with \texttt{OpenKP}, is the only neural method that outperforms TFIDF in this zero-shot evaluation. It outperforms previous neural methods by more than 60\%, and itself when trained on KP20k, confirming the strong generalization ability of \texttt{BLING-KPE} and the training with \texttt{OpenKP}.
\subsection{Discussion}

\begin{figure}
    \centering
    \includegraphics[width=0.95\columnwidth]{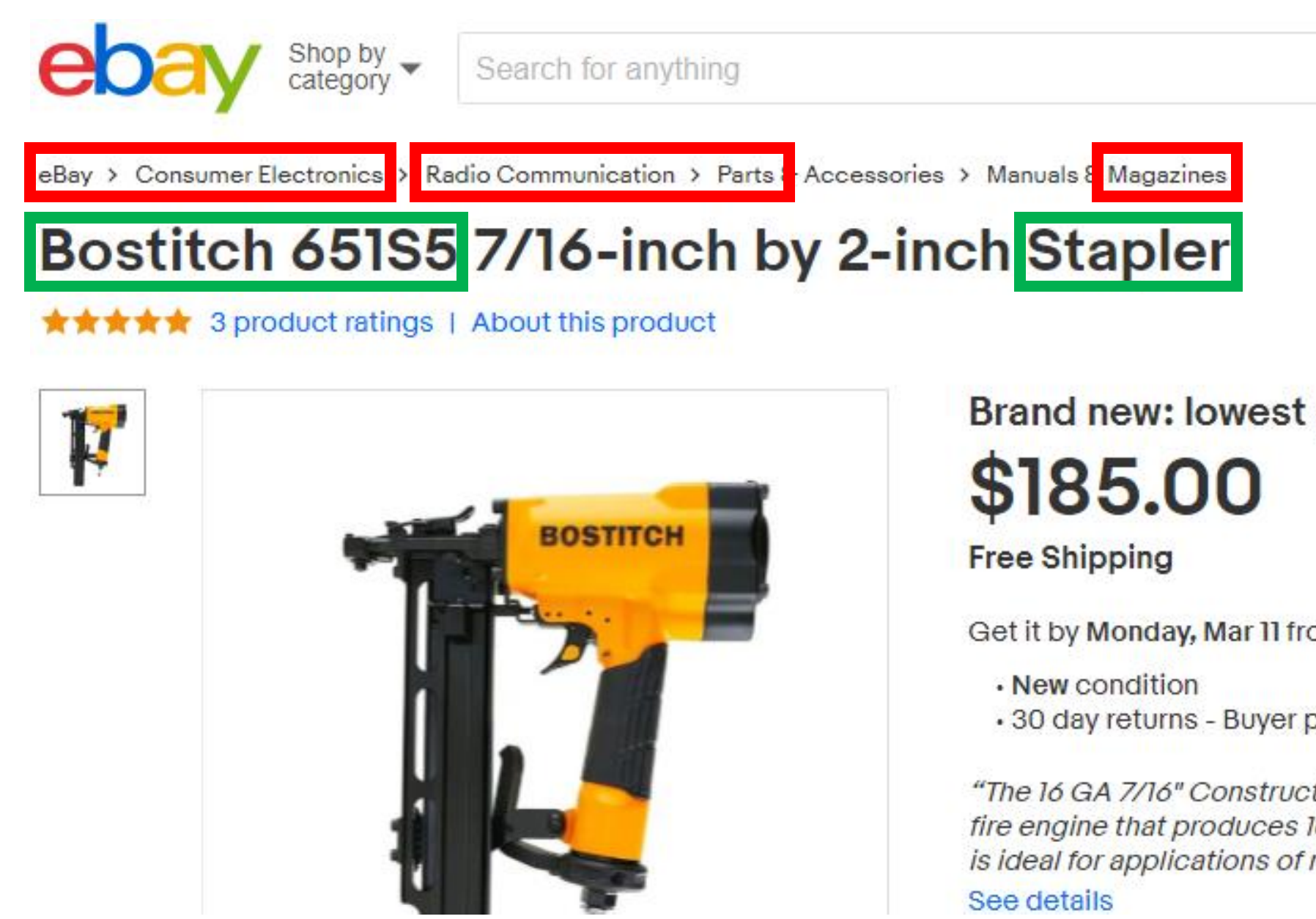}
    \caption{An example of visual feature contributions in BLING-KPE. The big green blocks are the keyphrases extracted with visual features. The small read blocks are those extracted without visual information.}
    \label{fig:eg}
\end{figure}

Our manual case studies found many interesting examples that illustrate the advantage of modeling documents with visual information. 

For example, in Figure~\ref{fig:eg}, the page is annotated with ``Bostitch 651S5'', the product name, and ``Stapler'', the product type. Their salience is highlighted by larger and bold fonts, which are picked up by BLING-KPE. However, without the visual information, the product ontology names are extracted as keyphrases: they are meaningful concepts, correlated with the page content, and positioned at the beginning of the document---only that they appear less important in the web page by design.

\section{Conclusion}
\label{sec:conclusion}






    
This paper curates OpenKP, the first public large scale open domain keyphrase extraction benchmark to facilitate future research keyphrase extraction research in real-world scenarios.
It also develops \texttt{BLING-KPE}, which leverages visual representation and search-based weak supervision to model real-world documents with variant contents, appearances, and diverse domains.

Our experiments demonstrate the robust improvements of BLING-KPE compared to previous approaches.
Our studies showcase how BLING-KPE's language understanding, visual features and search weak supervision jointly deliver this effective performance, as well as its generalization ability to an unseen domain in zero-shot setting. 

In the future, we plan to extend OpenKP with more annotated documents and connect it with downstream applications. 

\bibliographystyle{acl_natbib}
\bibliography{citation}
\flushend
\end{document}